# MFL-YOLO: An Object Detection Model for Damaged Traffic Signs

First A. TENGYANG CHEN, Second B. JIANGTAO REN

*Abstract*— **Traffic signs are important facilities to ensure traffic safety and smooth flow, but may be damaged due to many reasons, which poses a great safety hazard. Therefore, it is important to study a method to detect damaged traffic signs. Existing object detection techniques for damaged traffic signs are still absent. Since damaged traffic signs are closer in appearance to normal ones, it is difficult to capture the detailed local damage features of damaged traffic signs using traditional object detection methods. In this paper, we propose an improved object detection method based on YOLOv5s, namely MFL-YOLO (Mutual Feature Levels Loss enhanced YOLO). We designed a simple cross-level loss function so that each level of the model has its own role, which is beneficial for the model to be able to learn more diverse features and improve the fine granularity. The method can be applied as a plug-and-play module and it does not increase the structural complexity or the computational complexity while improving the accuracy. We also replaced the traditional convolution and CSP with the GSConv and VoVGSCSP in the neck of YOLOv5s to reduce the scale and computational complexity. Compared with YOLOv5s, our MFL-YOLO improves 4.3 and 5.1 in F1 scores and mAP, while reducing the FLOPs by 8.9%. The Grad-CAM heat map visualization shows that our model can better focus on the local details of the damaged traffic signs. In addition, we also conducted experiments on CCTSDB2021 and TT100K to further validate the generalization of our model.**

## I. INTRODUCTION

With the development of autonomous driving technology, the object detection of traffic signs has become an integral part of the autonomous driving field. The accurate detection and recognition of traffic signs is crucial for the safe driving of vehicles during the driving process of autonomous vehicles. Therefore, the application of object detection techniques for traffic signs in autonomous vehicles is also becoming increasingly important. There are already a number of mature algorithms in this area, which are divided into two main categories: traditional methods based on artificially designed features and deep learning methods based on convolutional neural networks. Traditional methods can be classified as: color-based, shape-based, and multi-feature fusion, where manually designed features extracted from candidate regions are fed into the classifier to obtain detection results. Using traditional methods requires significant labor costs and the manually designed methods do not always characterize all features well and have poor generalization capabilities in complex scenes. In contrast, deep learning methods learn features directly from a large amount of training data, using a rich convolutional hierarchy that can represent complex object features. These method have made significant improvements in traffic sign detection, such as improving small object detection performance [1], improving model robustness in complex weather scenarios [2], and lightweight improvements to achieve speed-accuracy trade-offs [3], etc.

Traffic signs are important facilities to ensure the safety and smooth flow of traffic, but due to various reasons such as natural factors, human factors and management factors, traffic signs may appear damaged as shown in **Fig. 1**. This not only reduces the reputation of road management, but also brings great safety hazards to drivers and pedestrians, so the detection of damaged traffic signs has become an issue that draws attention. However, there is a lack of research on the detection of damaged traffic signs. For the problem of detecting damaged traffic signs, we have tried two task paradigms: one is to use only damaged traffic sign as training samples to build a single classification object detection task, the other is using damaged traffic sign as one class and normal traffic sign as another class to build a binary classification object detection task. The first task paradigm only learns the features of traffic signs due to the lack of comparison with normal ones, which leads to the model easily detecting normal traffic signs by mistake in the actual detection process. The second task paradigm can learn the features of traffic signs and damage at the same time, and thus can better distinguish damaged traffic signs from normal ones, which greatly reduces the case of false detection. Therefore, we adopt the second task paradigm.

Data collection for damaged traffic signs suffers from a wide variety of traffic signs, large differences in different damage states and difficult data acquisition. If the coverage and quantity of data are insufficient, it may lead to leakage problems in the actual detection process, so this is a major challenge for data collection and pre-processing. At the same time, since damaged traffic signs are closer in appearance to normal intact

This work was supported in part by the Key R&D projects of Guangdong Province under Grant 2022B0101070002. *Corresponding author: Jiangtao Ren.*

First A. TENGYANG CHEN is with the School of Computer Science and Engineering, Sun Yat-Sen University, Guangzhou, 510275 China (e-mail: chenty26@mail2.sysu.edu.cn)

Second B, JIANGTAO REN is with the School of Computer Science and Engineering, Sun Yat-Sen University, Guangzhou, 510275 China (e-mail: issrjt@mail.sysu.edu.cn)

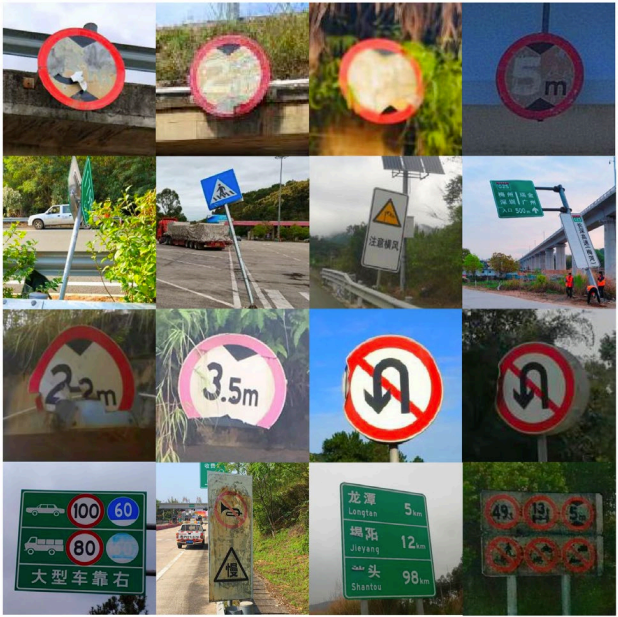

**Figure. 1** Some of the damaged traffic signs in our dataset.

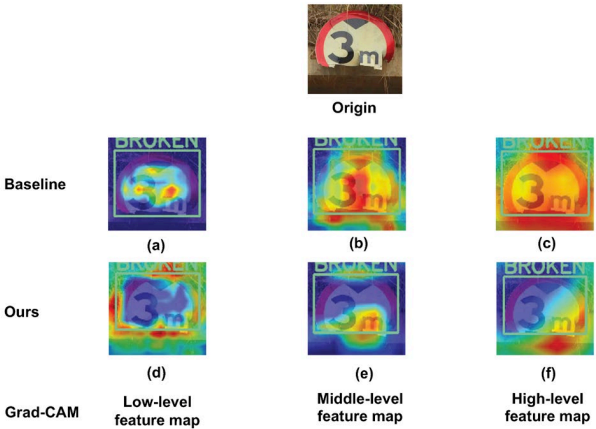

**Figure. 2** Comparison of grad-cam of YOLOv5s and MFL-YOLO's each level feature map.

traffic signs, and the damage states are complex and diverse, it may be difficult to capture the detailed local damage features in damaged traffic signs using current traffic sign detection methods to distinguish them from normal ones, resulting in a lower accuracy rate. Most of the existing object detection models, use a multi-level detection head design to capture information at different scales of the object and fuse features at different levels through FPN (Feature pyramid network), allowing the model to better learn features at different scales [4][5][6]. However, the existing improved feature FPN mainly considers how to fuse features at different levels so as to enhance the relevance of features at each level, but this may make the features learned by different detection heads closer to each other, which is not conducive to learning the abundant features at different levels. By visualizing the heat map of the feature maps of different detection heads in the baseline model, as shown in **Fig. 2**, we find that the focus of the low-, middle- and high-level of the baseline are all relatively close to each other, concentrating on the central area of the object and differing only in the size of the range, but the damaged areas of some traffic signs are mainly distributed at their edges, which is obviously not conducive to the detection of damaged traffic signs. Meanwhile, the task of detecting damaged traffic signs is often undertaken by mobile devices with limited computing power and storage space, so how to streamline the size of the model and reduce computational pressure while maintaining accuracy is also one of the issues to be considered.

To solve the problem above, we first constructed a damaged traffic sign dataset collected by Guangdong Leatoptech Investment Company, which contains 503 damaged traffic signs from urban roads, including 35 types of traffic sign types and diverse damage types such as worn, cracked, and tilted, and expanded the damaged traffic sign data to 2165 copies through data augmentation methods such as flipping, cropping, lightness and darkness, contrast adjustment, and compression and distortion. For the construction of normal traffic sign data, we introduced some data from the TT100K dataset [7] and the CCTSDB2021 dataset [8], and 2165 of them were manually filtered as the normal traffic sign class. Through data augmentation and manual filtering, we construct a well-balanced dataset, avoiding the problem of the imbalance of the real dataset. Inspired by the paper in fine-grained classification [9], we propose the damaged traffic sign detection algorithm MFL-YOLO based on YOLOv5s. We have concentrated on increasing the variability between high- and low-level feature maps, while also considering the interaction between low- and medium-level, and medium- and high-level feature maps. We designed $L_{MFL}$ (Mutual Feature Levels Loss), a loss function for different levels of feature maps, to achieve this. The loss function consists of a diversity component and a discriminative component: the diversity component first extracts the feature maps of each level through the convolution layer, and then calculates the similarity between each of the two after operations such as maxpooling. There are three detection heads in YOLOv5s, meaning that there are three different levels of feature maps, so our method will generate three similarities. The loss function obtained from the linear combination of these three similarities can influence the correlation and difference between these three levels of feature maps by controlling the coefficient before each similarity. When the coefficient is positive, the difference between the feature maps of the corresponding levels increases, when the coefficient is negative, the correlation between the feature maps of the corresponding levels increases. The discriminative component then extracts the feature maps of each object from each detection head, takes the feature maps of damaged traffic signs as positive samples and the feature maps of normal ones as negative samples, randomly selects positive and negative sample pairs, and subsequently performs cross-entropy calculations with the ground-truth labels after cross-channel maxpooling, global average pooling, and softmax on the positive and negative sample pairs respectively, which allows the feature maps of each level to contain as much discriminatory information as possible and enhances the differentiation between damaged and normal traffic signs. As shown in **Fig. 2**, the feature maps of each level in our model have better attention to localized damaged traffic signs.

To enable better embedding of the model into mobile devices, we have also made lightweight improvements to the model. The GSConv, GSBottleneck and VoVGSCSP modules are used in the neck part of YOLOv5s instead of the standard convolution and CSP modules [10], which reduces the size and FLOPs of the model while retaining high accuracy and makes the model more practical. Through experimental comparisons on our constructed dataset, our model achieves an mAP of 93.3%, a 5.1% improvement over baseline, and an F1 score of 90.0, a 4.3 improvement over baseline, while reducing FLOPs by 8.9%.

To summarize, the contributions of this paper are as follows.

(1) A novel and meaningful research question is proposed based on the existing traffic sign detection - damaged traffic sign detection. By automatically identifying damaged traffic signs, the process of manual inspection is eliminated, saving time and labor costs. By repairing or replacing the detected damaged traffic signs in time, the integrity of the signs can be ensured and the efficiency and safety of traffic can be improved. At the same time, this task has the difficulty of distinguishing damaged signs from normal ones due to their similarity with traditional object detection models, which is a special characteristic compared to other common object detection tasks, and therefore this research has important practical significance and application value.

(2) The proposed MFL-YOLO reduces the information overlap of traditional feature pyramids in feature fusion by designing a simple loss function to control the variability and interaction between the feature maps of each detection head, allowing the feature maps of different levels to focus on different regions of the object, and improving the representation of multi-level features to achieve the goal of capturing the subtle localized damaged features.

(3) We lightened the neck of baseline to reduce the size and FLOPs of the model while maintaining a high level of accuracy. The reduced resource consumption on the mobile side allows the mobile monitoring device to be better adapted to different hardware and network environments, facilitating the deployment of the model.

(4) We balanced positive and negative samples as much as possible by data augmentation and manual screening in our damaged traffic sign dataset.

## II. RELATED WORK

### A. Object Detection

In the field of object detection, many excellent methods have emerged, such as the RCNN series [11], YOLO series [12], and SSD [13]. These methods can usually be classified into two categories: two-stage detectors and one-stage detectors.

Two-stage detectors, such as RCNN [11], Fast RCNN [14], and Faster RCNN [15], usually first extract candidate boxes using algorithms such as Selective Search and EdgeBox, and then use classifiers and regressors to filter and refine these boxes, obtaining the final object detection results. The RCNN is one of the pioneering works in the field of object detection, which first proposed the combination of candidate box generation and convolutional neural networks. The Fast RCNN further improves the detection speed by introducing the RoI pooling layer. The Faster RCNN unifies the candidate box generation and classification regression modules by introducing the RPN network, greatly improving the detection speed and accuracy. These methods have high detection accuracy and stability, but their running speed is still relatively slow and not suitable for real-time detection scenarios.

One-stage detectors, such as the YOLO series, SSD, and RetinaNet [16], directly predict object bounding boxes and class probabilities from feature maps. Among them, the YOLO (You Only Look Once) series is undoubtedly one of the most popular models. YOLO was first proposed by Joseph Redmon et al. in 2016. Its main characteristics is the ability to quickly and accurately detect objects in images, whose speed is much faster compared to traditional two-stage detectors. Here is a brief introduction to the YOLO series of models: YOLOv1 [12] is the first version of the YOLO series, which is characterized by its fast speed but relatively weak object detection accuracy. YOLOv1 transforms the object detection problem into a regression problem, dividing an image into multiple grids, with each grid responsible for detecting an object and predicting its location and category. However, its performance is not ideal for small objects and densely populated scenes. YOLOv2 [17] is an improvement on YOLOv1. The biggest change is the addition of a multi-scale feature extraction module, which allows it to better handle small objects and densely populated scenes. Additionally, YOLOv2 introduces new techniques such as Batch Normalization layers and convolutional layers to replace fully connected layers, resulting in significant improvements in both performance and speed. YOLOv3 [18] improved mainly on YOLOv2 by adding new techniques such as multi-scale prediction and cross-level connections. Multi-scale prediction can handle objects of different sizes better, while cross-level connections combine features from both low and high-level to improve the accuracy of object detection. YOLOv4 [19] introduces various improvements, including the optimization of backbone networks, the enhancement of data augmentation methods, and the introduction of new loss functions, which enables better handling of large-scale datasets and high-resolution images. YOLOv5 [20] utilizes a lightweight network architecture, including CSPDarknet, SPP, and adopts the PANet structure. It also uses the Focus module to increase the receptive field of the model and reduce the number of parameters, improving the accuracy and speed of object detection. YOLOv6 [21] adopts the YOLOR network architecture and the CSPDarknet architecture of YOLOv5, while using the SENet module with attention mechanism to further improve the detection accuracy. YOLOv7 [22] adopts a more advanced Transformer architecture, as well as technologies such as channel attention mechanism, spatial attention mechanism, etc., making the model more generalizable and robust. YOLOv8 uses technologies such as hierarchical attention mechanism, grouped convolution, and applies adaptive convolution technology to the entire network, improving detection accuracy and speed.

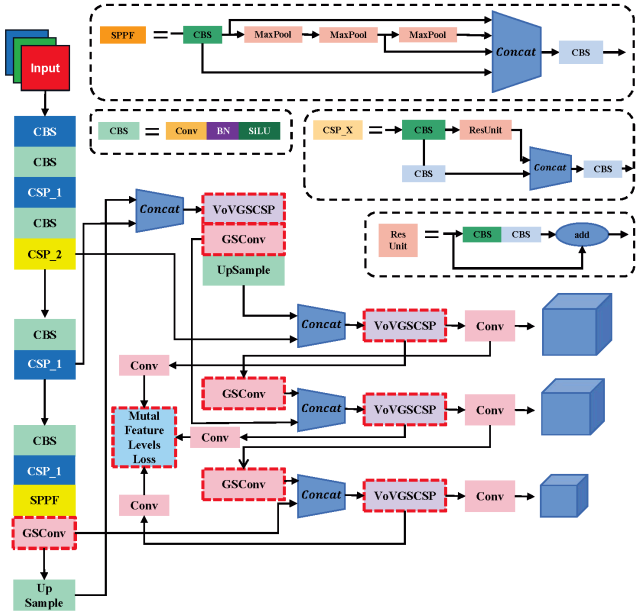

**Figure. 3** The network structure of MCL-YOLO. The specific structures of CBS, SPPF, CSP, and ResUnit are presented in the dotted boxes.

In addition, there are also some deep learning-based object detection methods, such as Mask RCNN [23] based on image segmentation, Cascade RCNN [24] based on regression and classification, etc. These methods add new features and characteristics to the original object detection algorithms, further improving detection performance and accuracy.

In summary, many excellent methods have emerged in the field of object detection, including both two-stage detectors and one-stage detectors, each with its own advantages and disadvantages. Future research can further explore how to improve detection accuracy, speed up detection, and adapt to more diverse scenarios based on these methods. Among the models mentioned above, YOLOv5s has achieved a good trade-off between speed, accuracy, and model size. Its strong generalization has been demonstrated in numerous custom datasets, and its open-source project has strong extensibility. Therefore, it has been widely used in various engineering tasks. In this experiment, we use YOLOv5s as our baseline, but its performance in distinguishing damaged and normal traffic signs is still not satisfactory, so corresponding improvements need to be made to the baseline model based on the characteristics of the data.

### B. Traffic Sign Detection

Traffic sign object detection is an important module in the perception layer of autonomous driving systems, which needs to accurately and quickly detect and recognize various traffic signs on the driving road. Due to the fact that the position of the traffic sign is not fixed, the color, shape, and size of the sign are diverse, and it is affected by factors such as lighting, occlusion, and blurring, making the task of traffic sign detection very complex.

Researchers have proposed various methods for traffic sign detection, which can be mainly divided into two categories: traditional image processing techniques and deep learning techniques. Traditional image processing-based methods [25] mainly use color, texture, and shape features for preprocessing, segmentation, and matching. These methods are simple and effective, but they are sensitive to parameter settings and have poor generalization ability. Methods based on deep learning techniques mainly use convolutional neural networks (CNNs) or their variants for end-to-end feature extraction and classification. These methods can automatically learn high-level semantic features and have higher accuracy and robustness. [3] proposed a fast and lightweight traffic sign detection algorithm called YOLOv5s-A2. The YOLOv5s network was improved by incorporating techniques such as depth wise separable convolution, attention mechanism, and anchor box clustering. The proposed algorithm achieves high detection accuracy while also improving detection speed and reducing model size. [1] proposed a model called SC-YOLO for small traffic sign detection, which improves YOLOv5 using techniques such as Cross Stage Partial Network (CSPN), Dense Connection Layer, and Spatial Information Preservation Layer. The model also introduces a dense neck structure and direction-aware SIOU to optimize the training process. SC-YOLO has fewer parameters and can improve the accuracy of small traffic sign detection.

The mentioned methods above mainly focus on detecting and recognizing normal traffic signs, and there is still a lack of research on detecting and recognizing damaged traffic signs. However, detecting damaged traffic signs is an important task for road maintenance work in the transportation department. Therefore, we hope that this paper can provide some help and bring some inspiring ideas for research in this field.

### III. METHODS

Our method is an improvement upon YOLOv5s, and the overall model structure is shown in **Fig. 3**. We add a Mutual Feature Levels Loss module to baseline, which contains a diversity component and a discriminative component. The diversity component controls the variability and interaction strength of each level by controlling the similarity between them. We decrease the similarity between high- and low-level feature maps and increase the similarity between middle- and high-level, low-level and middle-level feature maps by adjusting the positivity and negativity of the coefficients before the corresponding similarity so that they can each play their own role in capturing richer and finer-grained features. The discriminative component constructs positive and negative sample pairs and utilizes contrastive learning to make the learned features discriminative. We extract the feature maps of each level of the feature pyramid through the convolution layer and input them into the module for the calculation of the cross-level loss. The resulting loss will be linearly combined with the original classification loss function, localization loss function and confidence loss function in the baseline to guide the iterative optimization of the model through back propagation.

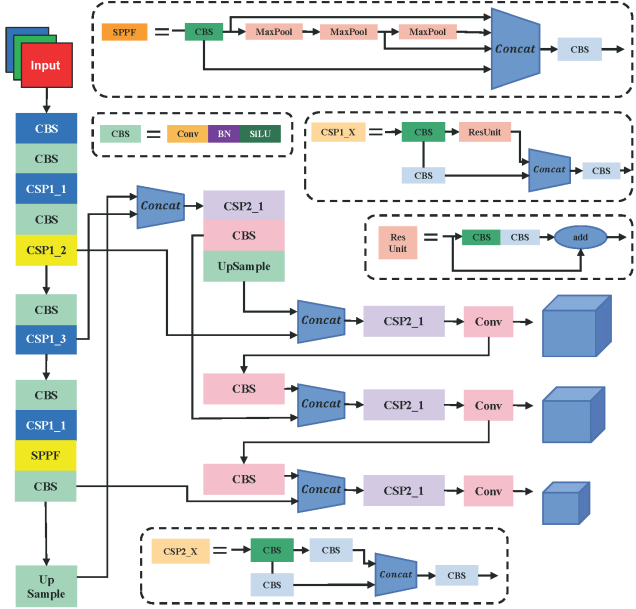

**Figure. 4** The network structure of YOLOv5s. The specific structures of CBS, SPPF, CSP, and ResUnit are presented in the dotted boxes.

We also conducted a lightweight transformation of the Neck part of YOLOv5s by using GSConv [10]. GSConv is a novel convolution operation that combines dense convolution with sparse convolution. In GSConv, some of the convolution kernels are fully connected while others are sparsely connected. This approach reduces the model parameters and computation while preserving the characteristics of dense convolution to enhance the model's expressiveness and accuracy. We replaced the standard convolution in the Neck part with GSConv and also replaced the standard convolution in the ResUnit and CSP modules within the baseline, adjusting their structures. We named the adjusted modules Ghostbottleneck and VoVGSCSP. The modified model has reduced parameters and computation while achieving improved accuracy. It is more suitable for deployment on mobile detection devices. We will introduce the details of each part next.

### A. YOLOv5

YOLOv5 is a one-stage object detection model that has achieved a good trade-off between accuracy and speed and has strong generalization on various datasets. Therefore, it is the most widely used version of the YOLO. YOLOv5 has four versions: YOLOv5s, YOLOv5m, YOLOv5l, and YOLOv5x. They share the same module and overall model structure, but differ in the width of channels and the depth of the network. Among them, YOLOv5s is the smallest version, and is selected as the baseline model for its fast detection speed and small model size, which are essential for the task of detecting damaged traffic signs.

The YOLOv5s model architecture consists of three main parts: the backbone, neck, and head, as shown in **Fig. 4**. The backbone uses CSPDarknet53 as the feature extraction network, which has fewer parameters and computations and can extract rich features. In the neck part, YOLOv5s uses the FPN and PAN structures, where FPN+PAN interacts with different level feature maps extracted from the main network (usually selecting three levels) across levels. FPN is a top-down feature pyramid that passes strong semantic features from higher-levels to lower-levels. However, the simple top-down FPN structure is insufficient for passing localization information between levels. Thus, PAN adds a bottom-up pyramid structure on top of FPN to pass localization information from lower-levels to higher-levels. FPN+PAN structure achieves cross-scale information fusion, making feature maps of each level have strong semantic and localization information, which is an effective method to improve the network's performance on multi-scale object detection tasks. In the head part, the model performs convolutional operations on each feature map to extract object features of different scales and predicts the position, size, and classification of target boxes.

The loss function of YOLOv5s mainly includes three aspects of loss: $L_{box}$ (the location loss), $L_{obj}$ (confidence loss) and $L_{cls}$ (classification loss). $L_{cls}$ is a cross-entropy loss function, which is used to calculate the difference between the predicted category and the ground truth. $L_{box}$ function is a GIoU loss function, which is used to calculate the difference between the predicted box and the true box; $L_{obj}$ is a binary cross-entropy loss function, which is used to calculate the difference between the probability that the predicted box contains an object and the probability that the true box contains an object.

In addition to the aforementioned components, YOLOv5s also employs multiple techniques to further enhance its performance. Firstly, the model applies Mosaic data augmentation, which is a method based on image splicing. During training, the model randomly selects four different images, splices them into a single large image, and then applies random cropping and scaling operations to obtain the training sample. This data augmentation technique can increase the diversity of training data and improve the model's robustness. Secondly, the Focal loss [16] is used to address the problem of sample imbalance by weighting the samples. In YOLOv5s, the model adjusts the weight of each sample based on the frequency of the object category, thereby making the model more focused on the detection performance of minority categories. Thirdly, the GIoU loss [26] is employed, which not only considers the degree of overlap between the target box and the ground-truth box but also takes into account the distance between them. By introducing the GIoU loss function, YOLOv5s can more accurately evaluate the quality of the object detection results. Finally, the model adopts the cosine annealing learning rate adjustment strategy [27], which is a method for addressing the problem of learning rate adjustment during neural network training. This strategy can prevent the learning rate from decreasing too quickly during training.

YOLOv5s integrates many of the state-of-the-art techniques and is an efficient and powerful object detection model. Through experimental comparisons, YOLOv5s has shown superior performance in detecting damaged traffic signs, which is why it is chosen as the baseline in our research.

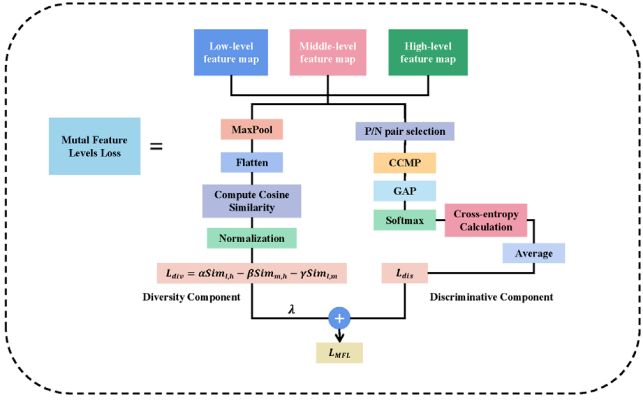

**Figure 5** The network structure of Mutual Feature Levels Loss. Where CCMP means cross-channel maxpooling.

### B. Mutual Feature Level Loss

This part is where our main innovation lies. One of the biggest challenges in detecting damaged traffic signs is that the damaged features are often local and subtle, and are usually distributed in the more peripheral parts of the sign, making it difficult for baseline models to capture these features. The reason may be that the FPN+PAN structure of YOLOv5s performs feature transfer and fusion between levels by up-sampling, down-sampling and concat operations, although it is simple and effective, this process also tends to cause partial information overlap and redundancy, which leads to some detailed part of information loss. The heat map visualization of the feature maps of each level shows that the features learned by the baseline model are mainly focused on the overall contour of the traffic signs as well as the object center (as shown in **Fig. 1**), which reflects a strong consistency. The high consistency between the feature maps of each level may be advantageous for coarse-grained object detection (detection task with large differences in appearance between different classes of objects), but is instead detrimental for fine-grained object detection like damaged traffic signs (detection tasks with subtle differences in appearance between different classes of objects). For coarse-grained object detection, sometimes only the general outline and color features of the objects are sufficient to distinguish it, while for fine-grained object detection, the model needs to extract enough rich and diverse discriminative features. Therefore, we want each level's feature map to learn as many different features as possible, in addition to all having strong semantic information as well as localization information, and to focus on the damaged local part of traffic signs as much as possible, so as to improve the model's ability to locate and distinguish damaged traffic signs. Unlike other studies that focus on making the feature maps of different levels interact and fuse better by designing more complex network structures [4][5][6], this paper does not make any changes to the original FPN and PAN structures, but achieves better results by designing a simple $L_{MFL}$ (Mutual Feature Levels Loss). The loss function consists of two main components: a diversity component and a discriminative component as shown in **Fig. 5**. The purpose of the diversity component is to take into account the interaction between the levels while pulling apart the similarity between the levels to drive them to learn different rich and diverse features. and the discriminative component is to perform object-level comparison learning on each level's feature map to guide the them to learn discriminative features. By adding an additional loss function only increase a small portion of computation in the training process, and does not increase the computational complexity in the inference process, which will lead to a reduction in inference speed. What's more, the $L_{MFL}$ can be applied as a plug-and-play module for target detection models like YOLO with multi-level detection heads. It is implemented as follows:

**Diversity component**: The Neck of YOLOv5s finally outputs three feature maps at different scales. In order to align these three feature maps and retain the most significant parts of each level feature map, we first design a specific maxpooling function. This function not only allows the feature map to be max-pooled, but also allows us to specify the shape of the output. Let the width and height of the original feature map be $w$ and $h$. We expect its output dimensions to be $w'$ and $h'$. Take the minimum value of $w/w'$ and $h/h'$ as the scaling factor $s$. Then, the width and height dimensions of the original feature map are down sampled to $(w/s, h/s)$ by the nearest linear interpolation, and finally the down sampled feature map is passed through a maximum pooling layer, and the kernel size is set to $(\frac{w}{s \times w'}, \frac{h}{s \times h'})$, the feature map can be maxpooled and output to the specified size at the same time. The purpose of max pooling is to extract the most prominent features in each pooling window by selecting the maximum value, while discarding other less important features. This reduces the dimensionality of the features, improves computational efficiency, and avoids overfitting. Additionally, max pooling can give the model a degree of translation invariance, as the most prominent features are always retained regardless of the object's position on the feature map, making the model more robust.

Next, the feature map after maxpooling and alignment is flattened along the spatial dimensions to reduce the dimensionality of the feature map, highlight the overall and interactions between features, and facilitate the next calculation steps. Then, pairwise cosine similarity calculation is performed on the flattened feature maps. Let the processed low-, mid-, and high-level feature maps be denoted as $F_l$, $F_m$ and $F_h$, The equation for calculating similarity here is (1).

$$Sim_{l,h} = \frac{1}{2} \times \left( \frac{F_l \times F_h}{|F_l| \times |F_h|} + 1 \right)$$

$$Sim_{l,m} = \frac{1}{2} \times \left( \frac{F_l \times F_m}{|F_l| \times |F_m|} + 1 \right)$$

$$Sim_{m,h} = \frac{1}{2} \times \left( \frac{F_m \times F_h}{|F_m| \times |F_h|} + 1 \right) \qquad (1)$$

The cosine similarity is used as the method to calculate the degree, and the similarity is mapped to [0, 1] by the operation of translation and deflation to avoid generating negative number and to facilitate the calculation, where a larger value means a higher similarity. The three values obtained represent the similarity between the three feature maps two by two, and subsequently we use the linear combination of these three similarities as the diversity loss function as shown in (2) to control the variability and interaction strength between each level of feature maps, where $\alpha$, $\beta$, $\gamma$ are hyperparameters.

$$L_{div} = \alpha Sim_{l,h} - \beta Sim_{l,m} - \gamma Sim_{m,h} \ (\alpha, \beta, \gamma > 0) \quad (2)$$

When the coefficient is positive, the similarity between the corresponding feature maps is reduced in the iterative process, and the difference between them increases, which is conducive to the extraction of rich and diverse features from the two feature maps. When the coefficient is negative, the similarity between the corresponding feature maps is increased, and the difference between them decreases, which is conducive to enhancing the information interaction between them. We chose to increase the difference between low- and high-level feature maps by setting the coefficient of $Sim_{l,h}$ to be positive. while increase the similarities between low- and middle-level, middle- and high-level feature maps by setting the coefficient of $Sim_{l,m}$ and $Sim_{m,h}$ to be negative.

Generally speaking, the low-level feature map mainly contains detailed information of the image, such as texture, edge. The middle-level feature map mainly contains some medium abstract information, such as the shape and outline of the object. The high-level feature map mainly contains higher-level semantic information, such as the category and location of the object. Increasing the difference between low-level and high-level feature maps can make them perform their respective roles better, capture as rich and diverse features as possible, and reduce in formation overlap. While the mid-level feature map acts as a bridge to transfer the information. Specifically, the mid-level feature map can transfer and integrate the information between the low-level feature map and the high-level feature map. Therefore, reducing the discrepancy between the low-level and mid-level feature maps and the discrepancy between the mid-level and high-level feature maps will be more conducive to information interaction, so that the mid-level feature map can better play the role of a bridge.

***Discriminative component***: Although the diversity module can make the feature maps of each level work separately and reduce the repetition of the learned features, simply pulling the differences of the feature maps of each level cannot ensure that the features are discriminative, but it may make the model learn some irrelevant features and lose the key information. Therefore, regularization constraints need to be imposed on the discrepancy pulling process to avoid model overfitting and to ensure that each level learns as many discriminative features as possible. We design a discriminative loss to achieve this.

First, for each object in a batch, the object-level feature map of each detection head is extracted and flattened in the spatial domain to obtain $Fo^{l,j}$, $Fo^{m,j}$ and $Fo^{l,j}$ ($l$, $m$ and $h$ represent low-, medium- and high-level respectively. $j = 0,1$. Which 0 represents damaged, while 1 represents normal) Let $N_0$ be the number of damaged objects in a batch. For each damaged traffic sign (positive sample), a normal traffic sign (negative sample) is randomly selected from each level to form $N_0$ positive-negative sample pairs.

Then, the extracted object-level feature maps are passed through a cross-channel max pooling operation, which maps the parts with maximum response within each channel of the feature map to a new feature map with a reduced dimensionality of 1 channel. Next, global average pooling is applied in the spatial domain. At this point, the object feature map is compressed to a scalar (3), where $W$ and $H$ represent the width and height of the feature map, and $N$ represents the number of channels.

$$g(Fo^{i,j}) = \frac{1}{WH} \sum_{k=1}^{WH} \max_{c=1,2...N} Fo_{c,k}^{i,j} \ (i = l, m, h) \quad (3)$$

Finally, a softmax is applied to each positive-negative sample pair and cross-entropy is calculated with the ground truth label. The average value of all sample pairs is taken as the discriminative loss function (4).

$$L_{dis} = \frac{\sum_{i=l,m,h} \sum_{j=1}^{N_0} L_{CE}\left(y, \frac{e^{g(Fo^{i,0})}, e^{g(Fo^{i,1})}}{e^{g(Fo^{i,0})} + e^{g(Fo^{i,1})}}\right)}{3 \times N_0} \quad (4)$$

The discriminative module combined with the diversity module can guide the model to capture diverse local discriminative regions, and the final $L_{MFL}$ is a linear combination of the two as shown in (5) , where $\lambda$ is hyperparameter.

$$L_{MFL} = L_{dis} + \lambda L_{div} \quad (5)$$

The $L_{MFL}$ will be linearly combined with the original $L_{cls}$ (the classification loss), $L_{box}$ (the localization loss) and $L_{obj}$ (the confidence loss) in the baseline to guide the iterative optimization of the model through back propagation as shown in (6).

$$L = L_{MFL} + L_{cls} + L_{box} + L_{obj} \quad (6)$$

### C. Slim Neck

To make the model more deployable on mobile devices, we also made lightweight modifications to the model. We used GSConv [10] to lightweight the neck part of YOLOv5s, which we call the Slim-Neck.

To speed up the computation during the model inference, the image goes through a similar transformation process: spatial information is gradually transmitted to channels, which can lead to the loss of some semantic information. Dense convolution computation can maximize the preservation of hidden connections between each channel, reducing the loss of semantic information. On the other hand, although sparse convolution speeds up the data transmission, it can also cut off these connections.

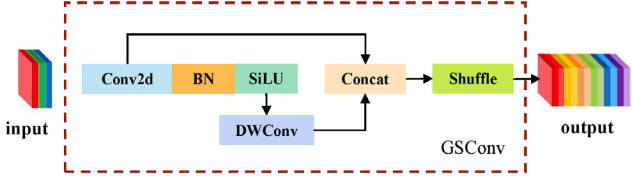

**Figure. 6** The network structure of GSConv. The "Conv" box consists of a convolutional-2D layer, a batch normalization-2D layer, and an activation layer. The "DWConv" means the DSC operation.

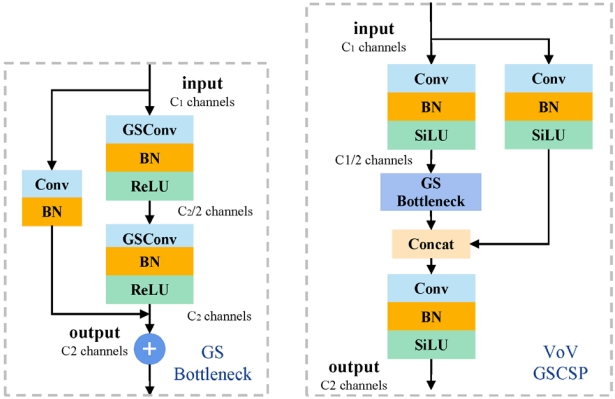

**Figure. 7** The network structure of GS Bottleneck and VoV-GSCSP.

GSConv is a novel convolutional operation that combines dense and sparse convolutions. Its specific structure is shown in **Fig. 6**. In addition to using L1 regularization to sparsify the convolutional kernel, GSConv also utilizes two techniques, Dynamic Receptive Field (DRF) and Half Dense Connection (HDC), to retain certain dense convolution characteristics. DRF technology can adaptively adjust the size of the convolutional kernel's receptive field during the convolution process to adapt to different image features. In GSConv, each convolutional kernel has a different receptive field size, which can improve the model's expression ability and accuracy. HDC technology introduces a certain proportion of fully connected operations in each group of convolutional operations to retain some dense connection characteristics. Specifically, in each group of convolutional kernels, some kernels are fully connected, while others are sparse connected. This can reduce the model's parameters and computational complexity while retaining certain dense convolution characteristics, thus improving the model's expression ability and accuracy.

To strike a trade-off between speed and accuracy, we utilized GSConv in the Neck part and replaced the original residual network structure in the CSP module of YOLOv5s with GSbottleneck. GSbottleneck consists of two GSConv layers and one standard convolution layer, with a BN layer and ReLU activation function added after each GSConv. The replaced CSP module is referred to as VoVGSCSP. The specific structure is shown in the **Fig. 7**.

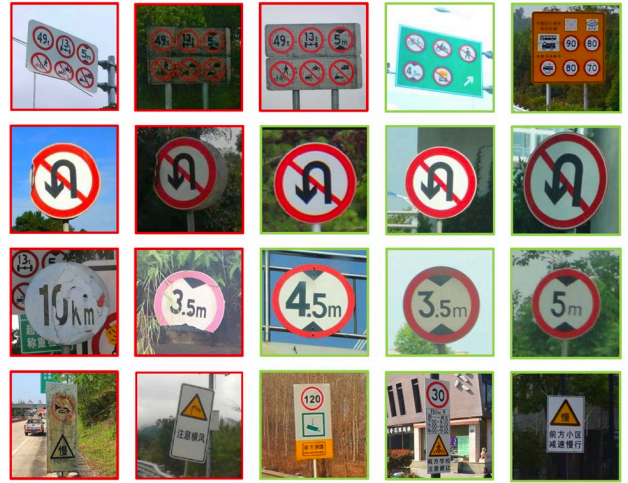

**Figure. 8** Some of the data in Dataset1, in order to better present the comparison of normal broken signs, the traffic signs in the original data are cropped out separately here. Where the red boxes indicate the damaged traffic signs and the green boxes indicate normal ones. In our dataset, the categories between damaged traffic signs and normal ones are aligned.

## IV. EXPERIMENTS

### A. Experimental Settings

1) **Datasets**

**Dataset1:** We constructed a damaged traffic sign dataset (as shown in **Fig. 8**), which contains 503 damaged traffic signs collected by Guangdong Leatoptech Investment Company, including 35 types of traffic signs and various broken states such as bent, worn, cracked and tilted, and we expanded the damaged traffic sign data to 2165 by flipping, cropping, brightness, contrast adjustment, sharpening and other data augmentation methods. The data were expanded to 2165. As for the normal traffic sign data. TT100K [7], CCTSDB2021 [8] and some normal traffic sign data collected online were introduced and some of the data were re-labeled to make the types of damaged traffic signs and normal ones align as much as possible. We manually filtered 2165 of the data as normal traffic signs.

**Dataset2:** CCTSDB2021 [8] is one of the most well-known traffic sign datasets in China, created by Changsha University of Science and Technology. The dataset contains three important types of traffic signs: "direction signs", "prohibition signs", and "warning signs", and includes six weather conditions, such as night, snow, and rain, which are more realistic and closer to real life. The training set contains 16,354 images, and the validation set contains 1,500 images.

**Dataset3:** TT100K [7] is a public dataset created by the School of Transportation Engineering at Tsinghua University in China in collaboration with Tencent's research lab. The dataset aims to provide a benchmark for research in autonomous driving, traffic monitoring, and intelligent transportation systems. The TT100K contains 151 categories, with 6,107 images in the training set and 3,073 images in the validation set. However, only 45 categories have more than 50 instances, and nearly half of the categories have only a single digit number of instances, leading to serious data imbalance issues. Therefore, we processed the dataset by only retaining categories with more than 50 instances in order to reduce the data imbalance problem.

2) **Experimental Configuration and Evaluation Indexes**
The experimental environment of this study was Windows 10 + CUDA 11.6 + Pytorch 1.12.0 + Python 3.7, and the experimental device was an NVIDIA GeForce RTX 3060 laptop GPU with 6 GB of memory. For the hyperparameter settings, the initial learning rate was set to 0.01, the learning rate strategy was cosine, the training cycle was 400, the initial input size of the model was 640×640, and the batch size was 16. We set $\alpha = 4$, $\beta = 0.5$, $\gamma = 0.5$ and $\lambda = 0.3$ in $L_{MFL}$. During model training, the mean average precision (mAP) and F1 score were chosen as evaluation metrics. A fixed intersection over union (IoU) value of 0.5 was chosen to calculate mAP in this study.

### B. Comparative Experiments

We conducted comparative experiments on Dataset1, Dataset2, and Dataset3. Among them, only Dataset1 contains damaged traffic signs, which is most closely related to the problem addressed in this paper. Therefore, we mainly conducted evaluation and analysis on this dataset. Dataset2 and Dataset3 both contain normal traffic signs, and the tasks constructed by these two datasets are to distinguish different types of normal signs. Although they are different from the problem we want to solve, they have some relevance. Therefore, we also conducted comparative experiments on these two datasets to verify the generalization of the model.

Due to the high requirements for real-time performance and accuracy in the task of detecting damaged traffic signs, we did not include the two-stage object detection models from the RCNN series and the outdated SSD in the comparative experiments. Instead, we chose the YOLO series models, which are widely used and more advanced, as the benchmark models for comparison. As shown in Table I, our method achieved the highest F1 score and mAP among these methods. The F1 score was improved by 4.3 and mAP was improved by 5.1 compared to the baseline. Thanks to our lightweight modification, the FLOPs was reduced by 1.4 compared to the baseline, and the model parameter increased only by 0.3M. Our method directly adjusts the interaction between the feature maps of each level in the FPN through the loss function, which is different from other methods that perform structural modification on FPN. We compared it with BiFPN [6], an FPN modification method with practical enhancement effect on Dataset1. The BiFPN network utilizes a bidirectional feature propagation approach to propagate and fuse features from different levels. Specifically, it achieves information propagation in two directions, top-down and bottom-up, and integrates features from different levels using a multi-level feature pyramid. The BiFPN network structure employs the idea of adaptive feature fusion and multi-level feature pyramid, which can effectively improve the accuracy and robustness of object detection. By using the BiFPN structure on the baseline, we observed improvements of 1.2 and 2.4 in F1 score and mAP, at the cost of a 0.6 increase in FLOPs and a 0.2M increase in model parameters. In contrast, our approach does not require any modification to the original FPN structure, making it easier to implement. Moreover, we achieved a greater improvement at a smaller cost.

We also conducted comparative experiments with different object detection methods on the CCTSDB2021 and TT100K datasets. As shown in Table II and Table III, our proposed method still achieved the best performance on these two datasets. Specifically, on the CCTSDB2021 dataset, compared with the baseline model, MFL-YOLO improved the F1 score and mAP by 1.3 and 1.4, respectively. On the TT100K dataset, compared with the baseline model, MFL-YOLO improved the F1 and mAP by 2.1 and 2.7, respectively. This further demonstrates the strong generalization ability of our object detection model in traffic scenes. However, due to the low granularity requirements of these tasks for the model, the improvement was not as significant on the Dataset1.

We used Grad-CAM to visualize the heat map for each

### TABLE I
THE DETECTION PERFORMANCE COMPARISON OF DIFFERENT METHODS ON DATASET1.

| Method | F1 | mAP | FLOPs | Params(M) |
|---|---|---|---|---|
| YOlOv3-tiny | 84.4 | 86.4 | 12.9 | 8.7 |
| YOlOv7-tiny | 75.0 | 76.2 | - | **6.0** |
| YOlOv8s | 83.1 | 7.5 | 28.4 | 11.1 |
| YOLOv5s(baseline) | 85.7 | 88.2 | 15.8 | 7.2 |
| YOLOv5s+BiFPN | 86.9 | 90.6 | 16.4 | 7.4 |
| MFL-YOLO(ours) | **90.0** | **93.3** | **14.4** | 7.5 |

### TABLE II
THE DETECTION PERFORMANCE COMPARISON OF DIFFERENT METHODS ON DATASET2 (CCTSDB2021).

| Method | F1 | mAP | FLOPs | Params(M) |
|---|---|---|---|---|
| YOlOv3-tiny | 81.8 | 81.2 | 12.9 | 8.7 |
| YOlOv7-tiny | 80.6 | 80.0 | - | **6.0** |
| YOLOv5s(baseline) | 81.2 | 80.9 | 15.8 | 7.2 |
| MFL-YOLO(ours) | **82.5** | **82.3** | **14.4** | 7.5 |

### TABLE III
THE DETECTION PERFORMANCE COMPARISON OF DIFFERENT METHODS ON DATASET3 (TT100K).

| Method | F1 | mAP | FLOPs | Params(M) |
|---|---|---|---|---|
| YOlOv3-tiny | 82.3 | 85.8 | 12.9 | 8.7 |
| YOlOv7-tiny | 82.0 | 83.9 | - | **6.0** |
| YOLOv5s(baseline) | 80.9 | 83.6 | 15.8 | 7.2 |
| MFL-YOLO(ours) | **83.0** | **86.3** | **14.4** | 7.5 |

TABLE IV
Ablation experiments of MFL-YOLO on Dataset1.

| Method | F1 | mAP | FLOPs | Params(M) |
|---|---|---|---|---|
| A: Baseline | 85.7 | 88.2 | 15.8 | 7.2 |
| B: SNC | 86.0 | 88.5 | 14.4 | **7.1** |
| C: SNC+Div | 90.0 | 92.8 | 14.4 | 7.5 |
| D:SNC+Div+Dis | **90.0** | **93.3** | **14.4** | 7.5 |

level of the baseline and MFL-YOLO, as shown in **Fig. 1**. It can be visually observed that the baseline model focuses on the center area of the object in each level of the feature maps, and the area of focus increases with the level of the feature maps, but the damaged area of the target is mainly distributed at the edge. In contrast, Our MFL-YOLO feature map effectively captures the damaged areas of the traffic sign well in all levels.

*C. Ablation Study*

To further analyze the effectiveness of our proposed individual modules, we performed ablation experiments on Dataset1. Since YOLOv5s has the best combined performance, we used YOLOv5s as the baseline." SNC" represents the slim neck, "Div" represents the diversity component in $L_{MFL}$, and "Dis" represents the discriminative component in $L_{MFL}$ , Experiment A represents the training results of baseline, neck, Experiment C represents the training results using slim-neck and diversity component, and Experiment D represents the training results using both slim-neck, diversity component and discriminative component (i.e., the complete $L_{MFL}$)

As shown in Table IV, each of the modules we proposed played a role in the model performance. Among them, the SNC module resulted in a 1.4 reduction in the FLOPs compared to baseline and a 0.1M reduction in the number of parameters, while F1 score improved by 0.3 and mAP improved by 0.3. The diversity component resulted in further improvements of 4.0 and 4.3 in F1 score and mAP, and finally the addition of the discriminative component resulted in a 0.5 improvement in mAP, proving the effectiveness of each module. Among them, since $L_{MFL}$ requires additional convolutional layers to extract the feature maps of each level during the training process, it increases the number of parameters of the model, but in fact these convolutional layers are not involved in the computation during the inference process and do not increase the computation of the model during the inference process.

V. Conclusion

In this paper, we improve on the mainstream object detection model YOLOv5s, so that the improved model can better capture local subtle features in damaged traffic signs which are different from normal ones, thus reducing the cases of confusing the two resulting in false detection. This is achieved by designing a loss function across multiple levels of feature maps, which consists of a diversity component and a discriminative component. The diversity component controls the diversity and interaction strength of each level by adjusting the similarity between them, so that each level can do its own job and capture as diverse and detailed features as possible, while the discriminative component exploits the idea of contrast learning to guide the diversity component with constraints to ensure that the captured features are highly discriminative. The method of controlling the interaction between the feature maps of each level by constructing a loss function does not require changes to the original model network structure, which is easy to implement, can be used as a plug-and-play module, and does not increase the computational effort of the model in the inference process. We also applied GSConv to lighten the model, reducing the number of parameters and computation, making it easier to deploy to mobile detection devices. We conducted comparative experiments on the dataset we constructed as well as the publicly available datasets CCTSDB2021 and TT100K to demonstrate the superiority and generality of our method. The ablation experiment part then verifies the effectiveness of each module. In our future work, we will focus on how to further compress the size of the model and improve the detection speed of the model at minimal cost, and hope that our work will stimulate more research on the problem of damaged traffic sign detection.

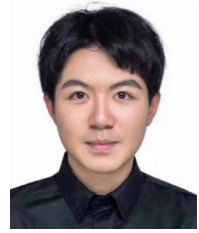

**First A. TENGYANG CHEN**, was born in Guangzhou, China, in 2000. He received the bachelor's degree in information & computing science from Sun Yat-Sen University. in 2018. He is currently pursuing the master degree in computer technology with Sun Yat-Sen University. His current research interests include applying neural networks in deep learning and image processing.

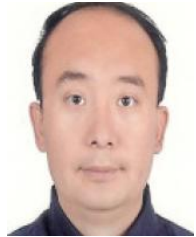

**Second B. JIANGTAO REN**, received the bachelor's degree in 1998 and the PhD degree in 2003 from Tsinghua University. He is currently an Associate Professor with the School of Computer Science and Engineering, Sun Yat-Sen University. His research interests include data mining and knowledge discovery, machine learning and intelligent transportation systems